\documentclass[runningheads]{llncs}

\usepackage[year=2026]{eccv}

\usepackage{eccvabbrv}

\usepackage{graphicx}
\usepackage{xcolor}
\newcommand{\legendbox}[2]{%
    \textcolor{#1}{\rule{0.8em}{0.4em}}~#2%
}
\usepackage{booktabs}
\usepackage{amsmath}
\usepackage{amssymb}
\usepackage{ulem}
\usepackage{ragged2e}

\usepackage[accsupp]{axessibility}  

\usepackage{hyperref}

\usepackage{orcidlink}

\begin{document}

\title{StateBridge: Phase- and Track-Grounded Reasoning for Synthetic-to-Real Traffic Video Captioning and VQA}
\titlerunning{StateBridge for Sim-to-Real Traffic Understanding}

\author{Blessing A. Kyem\inst{1,3} \and
Richard Dzinyela\inst{2} \and
Joshua K. Asamoah\inst{1,3} \and
Eugene Denteh\inst{1,3} \and
Andrews Danyo\inst{1,3} \and
Solomon Owusu-Ansah\inst{1,3} \and
Dickson G. Asiedu\inst{4} \and
Armstrong Aboah\inst{1,3}}

\authorrunning{B.A. Kyem et al.}

\institute{North Dakota State University, Fargo ND 58102, USA \and
Texas A\&M University, College Station 77840, TX, USA \and
Sustainable Mobility and Advanced Research in Transportation (SMART) Lab\and
Ghana Institute of Management and Public Administration}
\maketitle

\begin{figure}
    \centering
    \includegraphics[width=\textwidth]{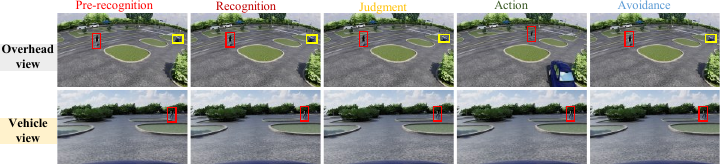}

    \vspace{0.3em}
    \begingroup
    \justifying
    \fontsize{7}{8.5}\selectfont

    \noindent
    \legendbox{red}{appearance}
    \hspace{0.8em}
    \legendbox{blue}{location}
    \hspace{0.8em}
    \legendbox{orange}{attention}
    \hspace{0.8em}
    \legendbox{green}{behavior}
    \hspace{0.8em}
    \legendbox{cyan}{context attributes}
    \par\vspace{0.3em}

    \noindent\textbf{Caption pedestrian:}
    \textcolor{red}{The pedestrian, a male in his 30s with a height of 170 cm, was wearing a black shirt and light blue jeans.}
    \textcolor{blue}{He was walking across a multi-lane road with two-way traffic.}
    \textcolor{blue}{The pedestrian's body was oriented perpendicular to the vehicle, and he had a moderate relative distance from it.}
    \textcolor{orange}{Despite the approaching vehicle, he seemed to be focused on crossing the road.}
    \textcolor{green}{He continued walking across the road without losing his balance or falling.}
    \textcolor{cyan}{The event took place on a clear day with bright lighting, dry road surface conditions, and light traffic volume. The road was level and made of asphalt. There were landscaped traffic islands and a pedestrian crossing near the pedestrian.}

    \par\vspace{0.3em}

    \noindent\textbf{Caption vehicle:}
    \textcolor{blue}{The vehicle was approaching the pedestrian from the right and was at a moderate distance from him.}
    \textcolor{blue}{Within the vehicle's field of view, the pedestrian was visible.}
    \textcolor{green}{The vehicle took action and avoided a collision by slowing down as the pedestrian crossed the road.}
    \textcolor{red}{The environmental conditions indicated that the pedestrian was a male in his 30s with a height of 170 cm. He was wearing a black shirt as upper body clothing and light blue jeans as lower body clothing.}
    \textcolor{cyan}{The weather was clear with bright lighting. The road surface conditions were dry and level, with asphalt as the road surface type. The traffic volume was light on this multi-lane road, which had two-way traffic. Landscaped traffic islands and a pedestrian crossing were also present.}

    \par\vspace{0.3em}
    \endgroup

    \caption{Multi-view SynWTS sample across five behavioral phases: pre-recognition, recognition, judgment, action, and avoidance. The top row shows the overhead view, and the bottom row shows the vehicle view. Red bounding boxes indicate the pedestrian, while yellow bounding boxes indicate the vehicle.}
    \label{fig:teaser}
    \vspace{-1.7em}
\end{figure}

\begin{abstract}
Traffic safety videos provide rich evidence for understanding interactions between pedestrians and vehicles, but synthetic-to-real evaluation makes this setting challenging because models must learn from simulated scenes and generalize to real-world traffic. In this work, we present StateBridge, a traffic video-language framework designed to operate under this synthetic-only training protocol. Our approach uses phase-aware clip construction, pedestrian and vehicle track summaries, and structured prompts to ground vision-language models in stable traffic evidence rather than appearance alone. We fine-tune public vision-language models on synthetic captioning and visual question answering examples, then combine their outputs through fact-locked caption fusion and evidence-diverse VQA consensus to improve consistency across actor roles, behavioral phases, and question types. StateBridge achieves an official S2 score of 55.4679 on AI City Challenge 2026 Track 2 using only synthetic training data. These results demonstrate the effectiveness of structured evidence transfer for traffic safety captioning and visual question answering.
\end{abstract}

\section{Introduction}
\label{sec:intro}

Traffic safety video understanding is an important step toward safer and more interpretable intelligent transportation systems~\cite{kong2024wts}. Unlike generic video captioning, this task does not ask a model to produce a short scene description; it requires the model to explain a safety-critical pedestrian-vehicle interaction over time, including the actors' locations, their attention, their motion, and the influence of the surrounding road environment on the event. Fig.~\ref{fig:teaser} illustrates this multi-view, phase-structured setting. AI City Challenge 2026 Track 2 increases the difficulty of this problem by restricting all adaptation to synthetic Digital Twin videos, while evaluation is conducted on real-world WTS traffic footage~\cite{aicitytrack2,Tang26AICity26}.

This synthetic-to-real setting introduces a structured but uneven domain gap because some aspects of the scene transfer more reliably than others. In particular, the Digital Twin data \cite{PatilAICity26} preserve road layout, camera viewpoint, event timing, actor tracks, and the annotation schema used to describe each scenario. These elements define the task semantics and can transfer from simulation to real video. However, the visual cues that reveal fine-grained human behavior, such as pose, gaze, gait, contact dynamics, illumination, clothing detail, and small objects, are far harder to simulate faithfully, so a model trained on this data can learn the correct output style from synthetic supervision while relying on appearance evidence that becomes unreliable at test time.

The challenge is amplified by the structured outputs required by the task. Each event is divided into behavioral phases, and each phase requires separate pedestrian- and vehicle-centric captions organized around location, attention, behavior, and context, while the same scenario is also evaluated through multiple-choice questions about attributes, geometry, visibility, action, and environment. These outputs are different projections of the same traffic state, so generating them independently can produce fluent captions that contradict VQA answers, disagree across actor roles, or alter a correct distance, speed, or phase-specific action. This suggests that synthetic-to-real traffic understanding should transfer structured evidence rather than simulator appearance. Although prior traffic VLM systems have improved high-resolution perception, spatial prompting, phase decomposition, and caption refinement in real-data settings~\cite{pervaiz2025trafficvila,wu2025trafficinternvl,phimsiri2025trafficinternvl,nguyennhu2025ster}, the synthetic-only protocol requires a more conservative strategy that exploits stable phase, role, geometry, and checklist semantics while limiting unsupported language changes caused by uncertain visual cues.

Building on this view, we propose \textbf{StateBridge}, a synthetic-only framework for phase- and track-grounded traffic safety captioning and VQA. StateBridge aligns each scenario by behavioral phase and actor role, summarizes pedestrian and vehicle tracks in a normalized coordinate system, and uses these signals as stable evidence for synthetic-trained video-language adapters. Caption hypotheses are fused through a fact-locking gate that accepts a role-phase edit only when it preserves supported attributes, relations, actions, and context. VQA hypotheses are combined through evidence-diverse consensus to reduce sensitivity to a single model, prompt, or grounding view. This design connects captioning and VQA through the same structured traffic evidence without retrieving target captions or adapting on real-world training videos.

Based on this formulation, this paper makes the following contributions:
\begin{enumerate}
    \item We formulate Track 2 as \textbf{asymmetric structured-state transfer}, where geometry, phase metadata, actor roles, and output semantics are more stable across domains than appearance and fine-grained human dynamics.
    \item We introduce a \textbf{phase- and track-grounded evidence interface} that aligns multi-view clips with pedestrian and vehicle trajectories and expresses actor motion through resolution-independent temporal box summaries.
    \item We propose \textbf{fact-locked caption fusion}, a conservative role-phase selection strategy that accepts caption edits only when they preserve supported attributes, relations, context, and phase-specific behavior.
    \item We develop an \textbf{evidence-diverse VQA consensus} over synthetic-trained adapters and provide a complete synthetic-only pipeline for dataset construction, model adaptation, inference, and official submission.
\end{enumerate}

\section{Related Work}
\subsection{Traffic Video-Language Understanding}
Video-language understanding has evolved from generic event description and question answering to safety-critical reasoning in road scenes. Early benchmarks such as MSVD, MSR-VTT, ActivityNet Captions, MovieQA, TGIF-QA, TVQA, and ActivityNet-QA established video captioning, temporal localization, and VideoQA as core tasks~\cite{chen2011msvd,xu2016msrvtt,krishna2017dense,tapaswi2016movieqa,jang2017tgifqa,lei2018tvqa,yu2019activitynetqa}, while recent VLMs and video-language models expanded these capabilities through multimodal instruction tuning, high-resolution perception, memory compression, token reduction, and long-context video reasoning~\cite{li2023blip2,dai2023instructblip,liu2023llava,qwen2025qwen25vl}. Driving-language datasets then brought these ideas closer to transportation by grounding commands, explanations, causal traffic reasoning, autonomous-driving VQA, graph reasoning, map-language understanding, and reliability analysis in road scenes~\cite{deruyttere2019talk2car,kim2018bddx,xu2021trafficqa,qian2024nuscenesqa,sima2024drivelm,cao2024maplm,li2025vladbench,xie2025drivebench,kong2025vlrdriver,driver_manuver,smartphone_roughness,bus_delays}. WTS and AI City Track 2 further specialize this direction to pedestrian-centered traffic safety with synchronized views, phase-level captions, actor boxes, and structured safety attributes~\cite{kong2024wts,wang2024aicity,xuan2024divide,shoman2024enhancing,pervaiz2025trafficvila,park2025trafficvila,wu2025trafficinternvl,phimsiri2025trafficinternvl,ha2025domainaware,kyem2025dualmodel,nguyennhu2025ster,querymotion,traffic_anomaly}. These works show the value of decomposition, view selection, high-resolution inputs, box-guided prompting, task-specific adaptation, and caption refinement, but they primarily study real-data WTS settings. StateBridge instead targets the stricter synthetic-to-real protocol by treating phase, role, geometry, and checklist semantics as the transferable interface.

\subsection{Grounded Multimodal Reasoning}
Grounded multimodal reasoning connects language to specific objects, regions, relations, and temporal moments rather than treating an image or video as a single global feature. Object detectors and trackers provide the spatial and temporal substrate for this connection~\cite{ren2015fasterrcnn,carion2020detr,zhu2021deformabledetr,zhang2022bytetrack,fisheye,helmet_violation,low_light,regiob_based,stargan_powerline}, while scene-graph and compositional VQA datasets show that visual understanding often depends on attributes and relations between entities~\cite{visualgenome2017,hudson2019gqa,prompt_guided,gazegnn}. Referring-expression and phrase-grounding methods further align text with localized regions through modular reasoning, joint image-text pretraining, and language-conditioned detection~\cite{kazemzadeh2014referitgame,yu2018mattnet,lu2019vilbert,kamath2021mdetr,li2022glip,minderer2022owlvit,liu2023groundingdino,cracksegmenter}. Recent multimodal models extend this grounding interface to coordinate-aware dialogue, region instruction tuning, visual marks, promptable segmentation, reasoning segmentation, pixel-level grounding, and open-ended visual task decoding~\cite{peng2023kosmos2,chen2023shikra,zhang2023gpt4roi,you2023ferret,yang2023setofmark,kirillov2023sam,zou2023seem,lai2024lisa,rasheed2024glamm,qwen3vl,gazesam,context_cracknet,pavesam}. In videos, temporal grounding methods localize query-relevant spans or spatio-temporal tubes~\cite{zhang2020vslnet,zhang2020twodtan,lei2021momentdetr,yang2022tubedetr,deepsegmenter}, which is especially relevant for traffic scenes where the same actor changes position and intent across phases. StateBridge follows this grounded reasoning line, but it uses the boxes and phase annotations provided by the challenge as a synthetic-to-real evidence interface, summarizing tracks into normalized temporal boxes instead of relying on a target-domain detector.

\subsection{Structured and Consistent Caption Generation}
Structured caption generation aims to produce fluent language while preserving the semantic fields that should appear in the output. Early image and video captioning models learned this mapping with recurrent decoders, attention, temporal structure, dense region descriptions, object-aware attention, transformer decoders, and vision-language pretraining~\cite{vinyals2015showandtell,xu2015showattend,venugopalan2015s2vt,yao2015temporal,johnson2016densecap,anderson2018bottomup,huang2019aoa,cornia2020m2,li2020oscar,zhang2021vinvl,li2022blip,li2023blip2,dai2023instructblip}. Since fluent captions can still contain unsupported objects or relations, later work studied metric-aligned training, semantic proposition evaluation, explicit detection, novel-object coverage, scene-graph guidance, hallucination analysis, and reference-free image-text scoring~\cite{papineni2002bleu,banerjee2005meteor,lin2004rouge,vedantam2015cider,anderson2016spice,rennie2017selfcritical,lu2018neuralbaby,wang2018objectcounts,agrawal2019nocaps,li2019pointing,yang2019sgae,wang2019scenegraphs,rohrbach2018hallucination,hessel2021clipscore}. Controllable captioning further separates content planning from realization through diverse generation, coarse-to-fine decoding, adversarial evaluators, retrieval support, and robustness-aware captioning~\cite{li2024evcap,li2024retrievalrobustness,pavecap}. This structure is central in WTS-style traffic description, where captions follow recurring appearance, location, environment, attention, and action fields and where prior AI City systems have used segment decomposition, rewrite modules, high-resolution VLMs, and fact checking~\cite{xuan2024divide,pervaiz2025trafficvila,park2025trafficvila,phimsiri2025trafficinternvl}. StateBridge follows this line but applies structure conservatively: it treats the phase-role checklist as a fact-preserving interface and accepts only caption fields that remain consistent with the shared traffic state.

\section{Task and Problem Setup}

\subsection{Task Definition}

AI City Challenge 2026 Track 2 evaluates traffic safety understanding under a synthetic-to-real protocol~\cite{aicitytrack2}. The system is trained on labeled SynWTS Digital Twin videos and tested on real WTS videos without real-world supervision. Test inputs include videos, phase metadata, questions, answer options, and challenge-provided visual annotations when available, but no target captions, target answers, or additional manual target annotations. The required outputs are phase-level pedestrian and vehicle captions, plus multiple-choice VQA answers. Captions describe location, attention, behavior, and context for each actor, while VQA covers attributes, environment, geometry, visibility, action, and vehicle state. The official S2 score combines caption metrics with VQA accuracy.

\subsection{Synthetic-to-Real Structured-State Transfer}

Let $\mathcal{D}_{s}$ denote the labeled synthetic source domain. A scenario contains synchronized videos $\mathbf{X}^{s}$, actor tracks $\mathbf{B}^{s}$, phase boundaries $\mathbf{P}$, phase-level role captions $\mathbf{Y}^{c}$, and VQA pairs $\mathbf{Y}^{q}$. At test time, the target domain provides real videos $\mathbf{X}^{t}$, phase metadata, questions, answer options, and challenge-provided visual annotations when available, but no caption or answer supervision. For phase $p$ and role $r\in\{\mathrm{ped},\mathrm{veh}\}$, we define the task-relevant traffic state
\begin{equation}
    \mathbf{z}_{p,r}
    =
    \bigl(\mathbf{z}^{\mathrm{loc}}_{p,r},
    \mathbf{z}^{\mathrm{att}}_{p,r},
    \mathbf{z}^{\mathrm{beh}}_{p,r},
    \mathbf{z}^{\mathrm{ctx}}_{p}
    \bigr),
    \label{eq:traffic_state}
\end{equation}
corresponding to location, attention, behavior, and shared context. Captioning verbalizes $\mathbf{z}_{p,r}$, while each VQA item queries one or more of its coordinates. Let $p_s(\mathbf{X}\mid\mathbf{z})$ and $p_t(\mathbf{X}\mid\mathbf{z})$ denote the source and target video distributions conditioned on the same traffic state, let $\mathbf{Y}$ denote the task outputs, and let $\mathcal{S}_s$ and $\mathcal{S}_t$ denote the source and target annotation schemas. The domain shift is asymmetric:
\begin{equation}
    p_s(\mathbf{X}\mid\mathbf{z})
    \neq
    p_t(\mathbf{X}\mid\mathbf{z}),
    \qquad
    \mathcal{S}_s(\mathbf{z},\mathbf{Y})
    =
    \mathcal{S}_t(\mathbf{z},\mathbf{Y}),
    \label{eq:asymmetric_gap}
\end{equation}
Equation~\ref{eq:asymmetric_gap} states that the rendering of a safety state changes across domains, but the meanings of phase, role, geometry, tracks, and checklist attributes remain stable. StateBridge therefore transfers structured evidence rather than simulator appearance: at inference time it estimates support for $\mathbf{z}$ from the official target inputs and constrains captions and VQA by that evidence, without using target labels, target captions, target answers, or real WTS training videos.

\section{Method}
\label{sec:method}

\subsection{Framework Overview}

\begin{figure}
    \centering
    \includegraphics[width=\textwidth]{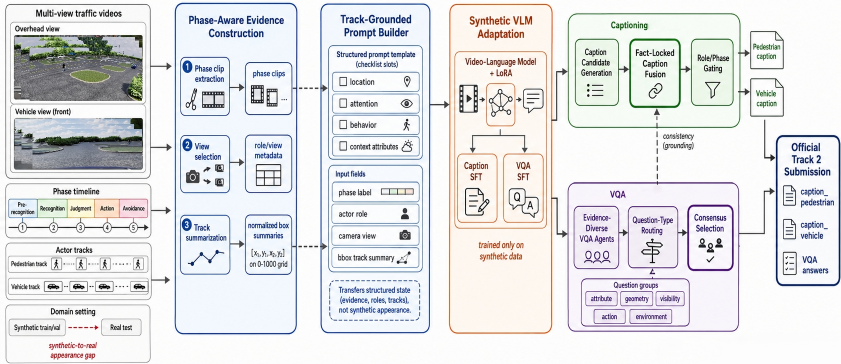}
    \caption{Overview of the proposed StateBridge framework.}
    \label{fig:framework}
\end{figure}

StateBridge converts each traffic scenario into a phase- and role-grounded evidence representation before generating captions or VQA answers. Fig.~\ref{fig:framework} summarizes the full pipeline from synthetic evidence construction to official caption and VQA outputs. This design follows the formulation in Eq.~\ref{eq:traffic_state}: instead of asking a VLM to transfer simulator appearance directly, we expose the stable elements of the event, namely phase, view, actor role, track geometry, and task schema. For a phase $p$ and request $u$ (a caption request or a VQA item), StateBridge constructs
\begin{equation}
    \mathbf{e}_{p,u}
    =
    \Phi\!\bigl(\mathbf{X}_{p}^{1:V},
    \widetilde{\mathbf{B}}_{p}^{\mathrm{ped}},
    \widetilde{\mathbf{B}}_{p}^{\mathrm{veh}},
    p,u
    \bigr),
    \label{eq:evidence}
\end{equation}
where $\mathbf{X}_{p}^{1:V}$ are the $V$ phase-aligned video views, $\widetilde{\mathbf{B}}$ denotes normalized pedestrian and vehicle track summaries, and $\Phi$ is the deterministic evidence-construction function. A set of $M$ synthetic-trained VLM adapters $\{g_m\}_{m=1}^{M}$ maps this evidence to candidate outputs. The final prediction is produced by a task-specific controller,
\begin{equation}
    \widehat{y}_{p,u}
    =
    \Psi_u\!\left(
    \{g_m(\mathbf{e}_{p,u})\}_{m=1}^{M},
    \widehat{\mathbf{z}}_{p}
    \right),
    \label{eq:controller}
\end{equation}
where $\Psi_u$ denotes the controller for request type $u$, and $\widehat{\mathbf{z}}_{p}$ denotes facts supported by prompts, tracks, and model outputs. The controller is deliberately asymmetric: free-form captions are selected through fact-preserving gates, while multiple-choice VQA answers are aggregated through evidence-diverse consensus.

\subsection{Phase-Aware Evidence Construction}

The first stage aligns every training and test example with the temporal unit used by the challenge. For each event and \texttt{normal\_trimmed} scenario, we build a manifest containing the split, scenario identifier, view, video path, phase interval, caption fields, VQA items, and available tracks. If phase $p$ spans timestamps $(s_p,e_p)$ and $\mathbf{X}^{v}$ is the full video from view $v$, each view is clipped as
\begin{equation}
    \mathbf{X}_{p}^{v}
    =
    \mathbf{X}^{v}[s_p:e_p],
    \qquad v=1,\ldots,V .
    \label{eq:phase_clip}
\end{equation}
where $V$ is the number of available views and $[s_p:e_p]$ denotes temporal clipping. Phase-specific captions and action or relation questions use these aligned clips, while environment questions retain broader scene context when the question is not tied to a short behavior interval. Each instruction states the scenario type, view, phase label, task, and expected output format. This construction makes the training instance match the evaluation instance and reduces leakage from neighboring behavioral phases.

\subsection{Track-Grounded Prompting}

Actor tracks provide a domain-stable cue because relative position and motion are less sensitive to rendering style than texture or illumination. Let $\mathbf{b}_{\rho}=[x_{1}^{\rho},y_{1}^{\rho},x_{2}^{\rho},y_{2}^{\rho}]$ be the pixel-coordinate box for a track summary position $\rho$ in a frame of width $W$ and height $H$. We convert each pedestrian and vehicle box to a shared $0$--$1000$ frame grid,
\begin{equation}
    \widetilde{\mathbf{b}}_{\rho}
    =
    1000
    \left[
    \frac{x_{1}^{\rho}}{W},\frac{y_{1}^{\rho}}{H},
    \frac{x_{2}^{\rho}}{W},\frac{y_{2}^{\rho}}{H}
    \right],
    \qquad
    \rho\in\{\mathrm{first},\mathrm{mid},\mathrm{last},\mathrm{mean}\}.
    \label{eq:bbox}
\end{equation}
For each actor and phase, the prompt contains the first, middle, last, and mean boxes, together with the actor role, view, phase, and number of tracked frames. This compact summary exposes coarse displacement, scale change, and relative geometry without depending on video resolution. We also construct optional rendered overlays and pedestrian-vehicle interaction crops. These views can make small actors easier to inspect, but they can also shift the visual distribution, so StateBridge treats them as alternative evidence streams rather than replacing the raw phase clips.

\subsection{Synthetic VLM Adaptation}

We adapt public multimodal instruction models with LoRA using LLaMA-Factory~\cite{llamafactory}. Our main backbone is Qwen3-VL-8B-Instruct~\cite{qwen3vl}. Each supervised row follows one of two explicit schemas. A caption row contains the phase-aligned videos, phase label, view name, track summaries, and an instruction to return a JSON object with \texttt{caption\_pedestrian} and \texttt{caption\_vehicle}. A VQA row contains the relevant videos, question, answer options, phase or environment scope, track summaries when available, and an instruction to return only the correct option letter.

We train three task mixtures from these rows. The caption-only adapter uses only caption rows, the VQA-only adapter uses only VQA rows, and the joint adapter mixes both tasks so that captioning and question answering share the same traffic vocabulary. We also train evidence variants with the same targets but different inputs: a text-box variant serializes normalized actor boxes in the prompt, an overlay variant renders colored actor tracks on the clips, a crop variant adds pedestrian-vehicle interaction crops, and a joint-visual variant combines raw clips with the derived grounding views. Each adapter minimizes the autoregressive supervised fine-tuning loss
\begin{equation}
    \mathcal{L}_{\mathrm{SFT}}
    =
    -\sum_{n=1}^{N}
    \sum_{j=1}^{|Y_n|}
    \log p_{\theta+\Delta\theta}
    \left(y_{n,j}\mid y_{n,<j},\mathbf{e}_n\right),
    \label{eq:sft}
\end{equation}
where $N$ is the number of supervised examples, $Y_n=(y_{n,1},\ldots,y_{n,|Y_n|})$ is the target token sequence for example $n$, and $\mathbf{e}_n$ is its evidence prompt. The parameters $\theta$ are frozen base-model weights, and $\Delta\theta$ is the trainable low-rank update. All examples in Eq.~\ref{eq:sft} come from the synthetic source domain; target videos are used only at inference.


\subsection{Fact-Locked Caption Fusion}

Caption candidates differ in useful but risky ways. One model may produce a better action phrase, while another may better preserve weather, road type, or distance. Because global rewriting can introduce unsupported attributes, we fuse captions at the role-phase field level. Let $c_0^{p,r}$ be the trusted field for phase $p$ and role $r$, and let $\mathcal{C}^{p,r}=\{c_k^{p,r}\}_{k=1}^{K}$ be alternative fields from other candidate systems. For each candidate, we compute
\begin{equation}
    S(c_k^{p,r})
    =
    \lambda_o O(c_k^{p,r},c_0^{p,r})
    + \lambda_n N_4(c_k^{p,r},c_0^{p,r})
    + \lambda_f F(c_k^{p,r},\widehat{\mathbf{z}}_{p})
    + \lambda_l L(c_k^{p,r},c_0^{p,r})
    - A(c_k^{p,r}),
    \label{eq:caption_score}
\end{equation}
where $O$ measures token overlap, $N_4$ measures four-gram preservation, $F$ rewards agreement with supported facts, $L$ rewards compatible length, $A$ penalizes repetition, contradictions, artifacts, and incomplete endings, and $\lambda_o,\lambda_n,\lambda_f,\lambda_l$ are fixed nonnegative weights. A candidate can replace the trusted field only when it improves the proxy score and passes a fact lock:
\begin{equation}
    S(c_k^{p,r})-S(c_0^{p,r})\geq\delta,\quad
    O(c_k^{p,r},c_0^{p,r})\geq\tau,\quad
    \mathrm{Lock}(c_k^{p,r},c_0^{p,r},\widehat{\mathbf{z}}_{p})=1.
    \label{eq:caption_gate}
\end{equation}
where $\delta$ is the required score margin, $\tau$ is the minimum token-overlap threshold, and $\mathrm{Lock}(\cdot)$ is a binary consistency test. In our implementation, $F$ is instantiated with checklist-term rewards for pedestrian, vehicle, weather, road, traffic, speed, and line-of-sight evidence; $A$ rejects incomplete endings, prompt artifacts, role contradictions, and impossible speed-collision combinations. The lock also protects identity, clothing, relative position, distance, obstacle, road, and weather terms when they are present. The final selected policy accepts fields whose length is between $0.84$ and $1.22$ times the trusted field and whose proxy score is no worse than $0.05$ below the trusted field, so style can improve only when the edit remains supported by the shared traffic state.

\subsection{Evidence-Diverse VQA Consensus}

VQA has a different error profile from captioning because the output space is a small set of option letters. Some questions are easiest from scene context, such as weather or road type, while others depend on phase-local geometry, visibility, or actor motion. We therefore combine adapters that see complementary evidence rather than relying on one prompt or one visual view. For a question $q$ with option set $\mathcal{A}(q)$, adapter $m$ predicts $\widehat{a}_{m}(q)$. The final answer is
\begin{equation}
    \widehat{a}(q)
    =
    \arg\max_{a\in\mathcal{A}(q)}
    \sum_{m=1}^{M}
    w_m\,
    \mathbb{1}\!\left[\widehat{a}_{m}(q)=a\right],
    \label{eq:vqa_vote}
\end{equation}
where $\mathbb{1}[\cdot]$ is the indicator function and $w_m$ is a fixed reliability weight for adapter $m$. The candidate pool contains synthetic-trained streams from the joint Qwen3-VL adapter, VQA-specialized Qwen3-VL adapters, and grounding variants built from text-box summaries, rendered box overlays, and interaction crops. In the final synthetic-only submission, we set $w_m=1$ for every included stream and use the strongest joint Qwen3-VL adapter as the tie fallback. This rule lets an overlay or crop model influence questions where it agrees with other evidence, but prevents a weaker grounding view from changing a high-confidence answer by itself.


\section{Experiments}

\subsection{Dataset and Protocol}

We train only on the official SynWTS synthetic train and validation splits. The preprocessing pipeline indexes event and \texttt{normal\_trimmed} scenarios, cuts phase-aligned clips, attaches available actor tracks, and exports instruction examples for captioning and VQA. This produces 14,198 supervised rows, including 2,465 phase-level caption examples and 11,733 VQA examples. At test time, the real WTS videos are used only for inference; no target labels, captions, answers, or manual annotations are added. We generate 798 caption rows and 19,624 VQA rows, then assemble them into the official scenario-level submission files.

We report the official server metrics: BLEU-4~\cite{papineni2002bleu}, METEOR~\cite{banerjee2005meteor}, ROUGE-L~\cite{lin2004rouge}, CIDEr~\cite{vedantam2015cider}, VQA accuracy, and the aggregate S2 score. Target captions and answers are hidden, so all model choices, thresholds, and fusion rules are fixed without target-label feedback.

\subsection{Implementation Details}

We train the main video-language adapters from Qwen3-VL-8B-Instruct using the Qwen3-VL chat template. Each adapter is optimized for three epochs with LoRA rank 64, LoRA scale 128, dropout 0.05, bfloat16 precision, gradient checkpointing, maximum sequence length 8,192, per-device batch size one, gradient accumulation of eight, learning rate $1\times10^{-4}$, cosine decay, and warmup ratio 0.03. The base model weights are frozen, and every prompt-target pair is generated automatically from SynWTS annotations.

We train three task mixtures. The caption-only adapter receives phase clips and predicts a JSON object with \texttt{caption\_pedestrian} and \texttt{caption\_vehicle}. The VQA-only adapter receives the question, answer options, relevant clips, and track evidence, and it is trained to output only the correct option letter. The joint adapter mixes both formats so that a single model learns the shared traffic vocabulary. We also train evidence variants that use the same targets but change the visual grounding: one uses only textual box summaries, one adds rendered track overlays, one adds pedestrian-vehicle interaction crops, and one combines raw clips with the visual grounding views.

For caption fusion, we assemble candidate captions from the Qwen3-VL adapters and additional synthetic-trained Qwen3.5-9B~\cite{qwen35_9b} and InternVL3-8B~\cite{internvl3} adapters. All candidates must be valid JSON with the two required role fields. If a candidate ends with an incomplete phrase, we apply a deterministic completion repair; if it drops a required role, changes the phase structure, or contradicts locked facts, it is rejected. For VQA, each candidate system produces one option letter per question. The final answer is selected by the equal-vote consensus in Eq.~\ref{eq:vqa_vote}, with the strongest joint Qwen3-VL adapter used only to break ties.

The systems compared in the tables differ only in how they combine these candidate outputs. The direct caption system uses one complete caption source without field replacement. Strict fact-locked fusion requires a larger score margin and length match, so it accepts fewer edits. Completion repair first removes incomplete final clauses before applying the same lock. Role-specific fusion applies separate pedestrian and vehicle edit budgets. Fixed-budget fusion accepts only the 45 or 48 highest-scoring locked fields. The final StateBridge system uses the selected fact-lock policy in Sec.~\ref{sec:method} without a fixed edit budget.

\subsection{Official Challenge Results}

Table~\ref{tab:bboard} reports official challenge-server results for the main synthetic-only submissions. All rows use the same evidence-diverse VQA consensus, so the differences isolate caption fusion behavior. The final StateBridge system obtains the best S2 among these verified variants by changing only fields that remain consistent with the supported traffic facts. Our approach ranks 8th in the Track 2 of the AI City Competition as shown in Table~\ref{tab:rank_snapshot}.

\begin{table*}[h!]
\centering
\caption{Official challenge-server results for synthetic-only submissions. Higher is better for all metrics.}
\label{tab:leaderboard}
\small
\resizebox{\linewidth}{!}{
\begin{tabular}{lcccccc}
\toprule
Method & S2 & BLEU-4 & METEOR & ROUGE-L & CIDEr & Acc. \\
\midrule
Direct caption + VQA consensus & 55.3330 & \textbf{0.2470} & 0.4199 & 0.4331 & \textbf{0.7502} & 81.2930 \\
Strict fact-locked fusion & 55.4614 & 0.2436 & 0.4244 & 0.4446 & 0.7243 & 81.2930 \\
Fact-locked fusion with completion repair & 55.4574 & 0.2436 & 0.4243 & 0.4446 & 0.7243 & 81.2930 \\
Role-specific fact-locked fusion & 55.4659 & 0.2438 & 0.4247 & 0.4446 & 0.7244 & 81.2930 \\
Fixed-budget fact-locked fusion, 45 fields & 55.4657 & 0.2438 & 0.4247 & 0.4446 & 0.7243 & 81.2930 \\
Fixed-budget fact-locked fusion, 48 fields & 55.4657 & 0.2438 & 0.4247 & 0.4446 & 0.7244 & 81.2930 \\
Rewrite model with fact locking & 55.4235 & 0.2423 & 0.4235 & 0.4432 & 0.7321 & 81.2930 \\
\midrule
\textbf{StateBridge: conservative fact-locked fusion} & \textbf{55.4679} & 0.2438 & \textbf{0.4248} & \textbf{0.4446} & 0.7252 & 81.2930 \\
\bottomrule
\end{tabular}
}
\end{table*}

\begin{table}[h!]
\centering
\caption{Official leaderboard ranking snapshot. Only rank, team ID, and aggregate score are shown.}
\label{tab:rank_snapshot}
\small
\begin{tabular}{ccc}
\toprule
Rank & Team ID & Score \\
\midrule
1 & 47 & 60.0853 \\
2 & 24 & 57.3307 \\
3 & 266 & 56.7949 \\
4 & 257 & 55.5901 \\
5 & 71 & 55.5768 \\
6 & 30 & 55.5728 \\
7 & 122 & 55.5123 \\
\textbf{8} & \textbf{127} & \textbf{55.4679} \\
9 & 176 & 55.4659 \\
10 & 97 & 55.3330 \\
\bottomrule
\end{tabular}
\end{table}

\subsection{Ablation Study}

Table~\ref{tab:ablations} summarizes verified submissions used to study the main components. The joint Qwen3-VL baseline uses one adapter for both captioning and VQA. The VQA consensus row keeps the direct caption source but replaces single-model VQA with the voting rule in Eq.~\ref{eq:vqa_vote}. The remaining rows compare increasingly constrained caption fusion strategies. These comparisons show that small, fact-locked edits are more reliable than broad rewriting or aggressive field replacement.

\begin{table*}[h!]
\centering
\caption{Ablation submissions for VQA consensus and caption fusion. Higher is better for all metrics.}
\label{tab:ablations}
\small
\resizebox{\linewidth}{!}{
\begin{tabular}{lcccccc}
\toprule
Variant & S2 & BLEU-4 & METEOR & ROUGE-L & CIDEr & Acc. \\
\midrule
Joint Qwen3-VL baseline & 55.0643 & 0.2442 & 0.4169 & 0.4284 & 0.6590 & 81.2424 \\
VQA consensus only & 55.3330 & \textbf{0.2470} & 0.4199 & 0.4331 & \textbf{0.7502} & 81.2930 \\
Rewrite model with fact locking & 55.4235 & 0.2423 & 0.4235 & 0.4432 & 0.7321 & 81.2930 \\
Fact-locked fusion with completion repair & 55.4574 & 0.2436 & 0.4243 & 0.4446 & 0.7243 & 81.2930 \\
Fixed-budget fact-locked fusion, 45 fields & 55.4657 & 0.2438 & 0.4247 & 0.4446 & 0.7243 & 81.2930 \\
Role-specific fact-locked fusion & 55.4659 & 0.2438 & 0.4247 & 0.4446 & 0.7244 & 81.2930 \\
\textbf{StateBridge: conservative fact-locked fusion} & \textbf{55.4679} & 0.2438 & \textbf{0.4248} & \textbf{0.4446} & 0.7252 & 81.2930 \\
\bottomrule
\end{tabular}
}
\end{table*}

\subsection{Qualitative Results}

\begin{figure*}[h!]
    \centering
    \includegraphics[width=\textwidth]{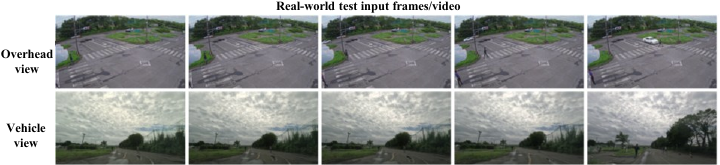}

    \vspace{0.3em}
    \begingroup
    \RaggedRight
    \fontsize{7}{8.5}\selectfont


    \noindent\textbf{Caption pedestrian:}
    \textcolor{red}{The pedestrian, a male in his 30s, stood still on a residential road. He was wearing a black T-shirt and black slacks, and his height was approximately 170 cm.}
    \textcolor{cyan}{The weather was cloudy, but the brightness was bright. The road surface was dry and level, made of asphalt. The traffic volume was light, and there were two-way lanes on the road.}
    \textcolor{blue}{The pedestrian's body was oriented in the opposite direction to the vehicle, which was directly in front of him. He had a close distance from the vehicle and had a clear line of sight in front, in the direction of travel.}
    \textcolor{orange}{The pedestrian was closely watching his surroundings, unaware of the vehicle's presence.}
    \textcolor{green}{He was not moving and appeared to be stationary.}

    \par\vspace{0.3em}
    \noindent\textbf{Caption vehicle:}
    \textcolor{blue}{The vehicle is positioned in front of the pedestrian, close in distance.}
    \textcolor{orange}{The pedestrian is visible within the vehicle's field of view.}
    \textcolor{green}{The vehicle is going straight ahead at a speed of 10 km/h.}
    \textcolor{red}{The environment conditions surrounding the event include a male pedestrian in his 30s, standing at a height of 170 cm. He is wearing a black T-shirt and black slacks.}
    \textcolor{cyan}{The weather is cloudy with bright brightness. The road surface conditions are dry and level, with asphalt as the road surface type. The traffic volume is light on this residential road with two-way traffic.}

    \par\vspace{0.35em}
    \noindent\textbf{VQA examples:}
    \par\vspace{0.15em}
    \noindent
    \begin{minipage}[t]{0.485\textwidth}
\textit{Phase: action.}\par
\textbf{Question:} What is the orientation of the pedestrian's body?\par
\textbf{Options:} a. Perpendicular to the vehicle and to the left; b. Opposite direction to the vehicle; c. Diagonally to the right, in the same direction as the vehicle; d. Diagonally to the right, in the opposite direction to the vehicle\par
\textcolor{green}{\textbf{StateBridge answer:} b. Opposite direction to the vehicle}
\end{minipage}\hfill
\begin{minipage}[t]{0.485\textwidth}
\textit{Phase: action.}\par
\textbf{Question:} What is the position of the pedestrian relative to the vehicle?\par
\textbf{Options:} a. On the left of the vehicle; b. Diagonally to the right, in front of the vehicle; c. Directly in front of the vehicle; d. On the right, behind the vehicle\par
\textcolor{green}{\textbf{StateBridge answer:} c. Directly in front of the vehicle}
\end{minipage}

    \endgroup

    \caption{Real public-test qualitative predictions from scenario \texttt{20230922\_7\_SN1\_T1}. The top row shows an overhead view and the bottom row shows the vehicle view across the five behavioral phases. The captions and VQA answers are StateBridge predictions produced at inference time.}
    \label{fig:real_test_qualitative}
\end{figure*}

Fig.~\ref{fig:real_test_qualitative} shows StateBridge predictions on a real public-test sequence. The model produces role-specific captions and VQA answers across the behavioral phases while using only synthetic training data. Fig.~\ref{fig:caption_repair} illustrates the caption gate: an incomplete direct output is repaired by removing the dangling phrase while preserving supported facts. Fig.~\ref{fig:vqa_consensus} illustrates the VQA consensus rule on a validation question, where complementary evidence streams are combined before the final option is selected. Together, the examples show that StateBridge relies on phase, role, track geometry, and checklist semantics rather than unconstrained text generation.

\begin{figure*}[h!]
    \centering
    \begingroup
    \setlength{\fboxsep}{4pt}
    \fontsize{7}{8.5}\selectfont
    \RaggedRight

    \noindent
    \begin{minipage}[t]{0.49\textwidth}
    \fcolorbox{blue!45!black}{blue!3}{%
    \begin{minipage}[t]{0.90\linewidth}
        \textbf{\textcolor{blue!55!black}{Direct VLM output}}\par
        \vspace{0.25em}
        \textit{Scenario \texttt{20230707\_11\_SY3\_T1}, phase: judgment, role: vehicle.}\par
        \vspace{0.45em}
        The vehicle is positioned diagonally to the left in front of the pedestrian, at a near distance.
        The pedestrian is visible within the vehicle's field of view.
        The vehicle is currently stopped, with a speed of 0 km/h.
        The environment conditions indicate that the pedestrian is a male in his 20s, with a height of 170 cm.
        He is wearing a black T-shirt and navy blue slacks.
        The weather is cloudy, with dim brightness.
        The road surface is dry and level, made of asphalt.
        The traffic volume is light on this residential road with two-way traffic.
        \textcolor{red}{There is an obstacle diagonally to the right}
        \par\vspace{0.35em}
        \textcolor{red}{\textbf{Issue:}} incomplete dangling phrase.
    \end{minipage}}
    \end{minipage}
    \hfill
    \begin{minipage}[t]{0.49\textwidth}
    \fcolorbox{green!45!black}{green!3}{%
    \begin{minipage}[t]{0.90\linewidth}
        \textbf{\textcolor{green!45!black}{StateBridge fact-locked repair}}\par
        \vspace{0.25em}
        \textit{Same scenario, phase, and role.}\par
        \vspace{0.45em}
        \textcolor{blue}{The vehicle is positioned diagonally to the left in front of the pedestrian, at a near distance.}
        \textcolor{orange}{The pedestrian is visible within the vehicle's field of view.}
        \textcolor{green!45!black}{The vehicle is currently stopped, with a speed of 0 km/h.}
        \textcolor{red}{The environment conditions indicate that the pedestrian is a male in his 20s, with a height of 170 cm. He is wearing a black T-shirt and navy blue slacks.}
        \textcolor{cyan!70!black}{The weather is cloudy, with dim brightness. The road surface is dry and level, made of asphalt. The traffic volume is light on this residential road with two-way traffic.}
        \par\vspace{0.35em}
        \textcolor{green!45!black}{\textbf{Repair:}} preserves supported facts and removes the incomplete ending.
    \end{minipage}}
    \end{minipage}


    \endgroup
    \caption{Qualitative example of fact-locked caption refinement. A direct VLM output preserves most traffic facts but ends with an incomplete phrase, while StateBridge keeps the supported geometry, visibility, speed, pedestrian appearance, and road context and removes the dangling ending without freely rewriting the event.}
    \label{fig:caption_repair}
\end{figure*}

\begin{figure*}[h!]
    \centering
    \begingroup
    \setlength{\fboxsep}{5pt}
    \fontsize{7}{8.5}\selectfont
    \RaggedRight

    \noindent
    \fcolorbox{black!20}{black!3}{%
    \begin{minipage}{0.965\textwidth}
        \textbf{Synthetic validation question.}
        Scenario \texttt{20230707\_18\_CY23\_T1}, phase: avoidance, scope: overhead view.
        \par\vspace{0.25em}
        \textbf{Question:} What is the pedestrian's awareness regarding vehicle?
        \par
        \textbf{Options:}
        a. Cannot be determined;
        b. Unaware of the vehicle;
        c. Notices the vehicle.
        \par\vspace{0.25em}
        \textcolor{green!45!black}{\textbf{Ground truth:} c. Notices the vehicle.}
    \end{minipage}}

    \vspace{0.55em}
    \noindent
    \begin{minipage}[t]{0.238\textwidth}
        \fcolorbox{green!45!black}{green!4}{%
        \begin{minipage}[t]{0.88\linewidth}
            \textbf{Phase clip + track text}\par
            Phase-local video with normalized pedestrian and vehicle track summaries.\par
            \vspace{0.35em}
            \textcolor{green!45!black}{\textbf{Predicts: c} \(\checkmark\)}
        \end{minipage}}
    \end{minipage}
    \hfill
    \begin{minipage}[t]{0.238\textwidth}
        \fcolorbox{green!45!black}{green!4}{%
        \begin{minipage}[t]{0.88\linewidth}
            \textbf{Joint caption--VQA VLM}\par
            Shared caption and VQA adaptation encourages answer consistency with traffic-state language.\par
            \vspace{0.35em}
            \textcolor{green!45!black}{\textbf{Predicts: c} \(\checkmark\)}
        \end{minipage}}
    \end{minipage}
    \hfill
    \begin{minipage}[t]{0.238\textwidth}
        \fcolorbox{orange!65!black}{orange!5}{%
        \begin{minipage}[t]{0.88\linewidth}
            \textbf{Visual grounding VQA}\par
            Overlay and crop evidence emphasizes local actor geometry but may miss awareness cues.\par
            \vspace{0.35em}
            \textcolor{orange!75!black}{\textbf{Predicts: b} \(\times\)}
        \end{minipage}}
    \end{minipage}
    \hfill
    \begin{minipage}[t]{0.238\textwidth}
        \fcolorbox{green!45!black}{green!4}{%
        \begin{minipage}[t]{0.88\linewidth}
            \textbf{VQA-only adapter}\par
            Task-specific adaptation focuses on option-letter prediction from phase evidence.\par
            \vspace{0.35em}
            \textcolor{green!45!black}{\textbf{Predicts: c} \(\checkmark\)}
        \end{minipage}}
    \end{minipage}

    \vspace{0.55em}
    \noindent
    \fcolorbox{green!45!black}{green!6}{%
    \begin{minipage}{0.965\textwidth}
        \textbf{\textcolor{green!45!black}{StateBridge evidence-diverse consensus:}}
        c. Notices the vehicle.
        The final answer is selected because it receives the strongest agreement across complementary evidence constructions, while the isolated visual-grounding disagreement is not sufficient to override the phase and track-supported answer.
    \end{minipage}}

    \endgroup
    \caption{Synthetic-validation illustration of evidence-diverse VQA consensus. The individual predictors are trained on synthetic data but receive different evidence constructions, including phase clips, textual track summaries, visual grounding views, and joint caption--VQA adaptation. Ground truth is shown only because this is a validation example.}
    \label{fig:vqa_consensus}
\end{figure*}

\subsection{Discussion}
The results show that the main difficulty is not only generating fluent text, but deciding when a generated field should be trusted. In Table~\ref{tab:leaderboard}, fact-locked fusion improves METEOR and ROUGE-L over the direct caption source, although it can reduce CIDEr when an accepted edit changes frequent reference-style phrases. This tradeoff explains why StateBridge uses field-level replacement rather than full-caption rewriting. The VQA consensus is similarly conservative: it changes answers only when complementary synthetic-trained streams provide enough support. These observations support the central design choice of StateBridge, namely to transfer phase, role, geometry, and checklist structure across domains while limiting unsupported changes in the final submission.

\section{Conclusion}
In this work, we present StateBridge, a synthetic-only framework designed to address the domain gap in traffic safety captioning and visual question answering, where models are trained on Digital Twin videos but evaluated on real-world traffic scenes. Our approach combines three core components that operate together: phase-aware evidence construction, track-grounded VLM adaptation, and conservative output fusion through fact-locked caption selection and evidence-diverse VQA consensus.

When evaluated on AI City Challenge 2026 Track 2, StateBridge achieved an official S2 score of 55.4679 under the synthetic-to-real protocol. These results show the value of transferring stable traffic structure, including phase, role, geometry, and checklist semantics, rather than relying only on appearance-level adaptation. Overall, StateBridge provides a practical foundation for traffic video-language understanding when real-world supervision is unavailable.


%
%
\clearpage
\bibliographystyle{splncs04}
\bibliography{main}
\end{document}